\pdfoutput=1
\documentclass[11pt]{article}

\usepackage{naacl2021}

\usepackage{times}
\usepackage{latexsym}

\usepackage[T1]{fontenc}
\usepackage[utf8]{inputenc}
\usepackage{graphicx}
\usepackage{multirow}
\usepackage{comment}
\usepackage{amsmath}
\usepackage{url}
\usepackage{microtype}

\title{Structure-Aware Abstractive Conversation Summarization\\ via Discourse and Action Graphs}

\author{Jiaao Chen \\
  School of Interactive Computing \\
  Georgia Institute of Technology \\
  \texttt{jiaaochen@gatech.edu} \\\And
  Diyi Yang \\
  School of Interactive Computing \\
  Georgia Institute of Technology \\
  \texttt{dyang888@gatech.edu} \\}

\begin{document}
\maketitle

\begin{abstract}
Abstractive conversation summarization has received much attention recently. However, these generated summaries often suffer from insufficient, redundant, or incorrect content, largely due to the unstructured and complex characteristics of human-human interactions. To this end, we propose to explicitly model the rich structures in conversations for more precise and accurate conversation summarization, by first incorporating discourse relations between utterances and action triples (``\textsc{who-doing-what}'') in utterances through structured graphs to better encode conversations, and then designing a multi-granularity decoder to generate summaries by combining all levels of information. Experiments show that our proposed models outperform state-of-the-art methods and generalize well in other domains in terms of both automatic evaluations and human judgments. We have publicly released our code at \url{https://github.com/GT-SALT/Structure-Aware-BART}. 
% \diyi{given that your abstract and title both mention "faithful", Related Work ans Evaluation should both highlight this as well}

\end{abstract}

% Document
% Intrinsic (synthesizing content using information in document) & Extrinsic (generate out of document) \cite{maynez-etal-2020-faithfulness}

\section{Introduction}
\begin{figure}[t]
\centering
\includegraphics[width=0.8\columnwidth]{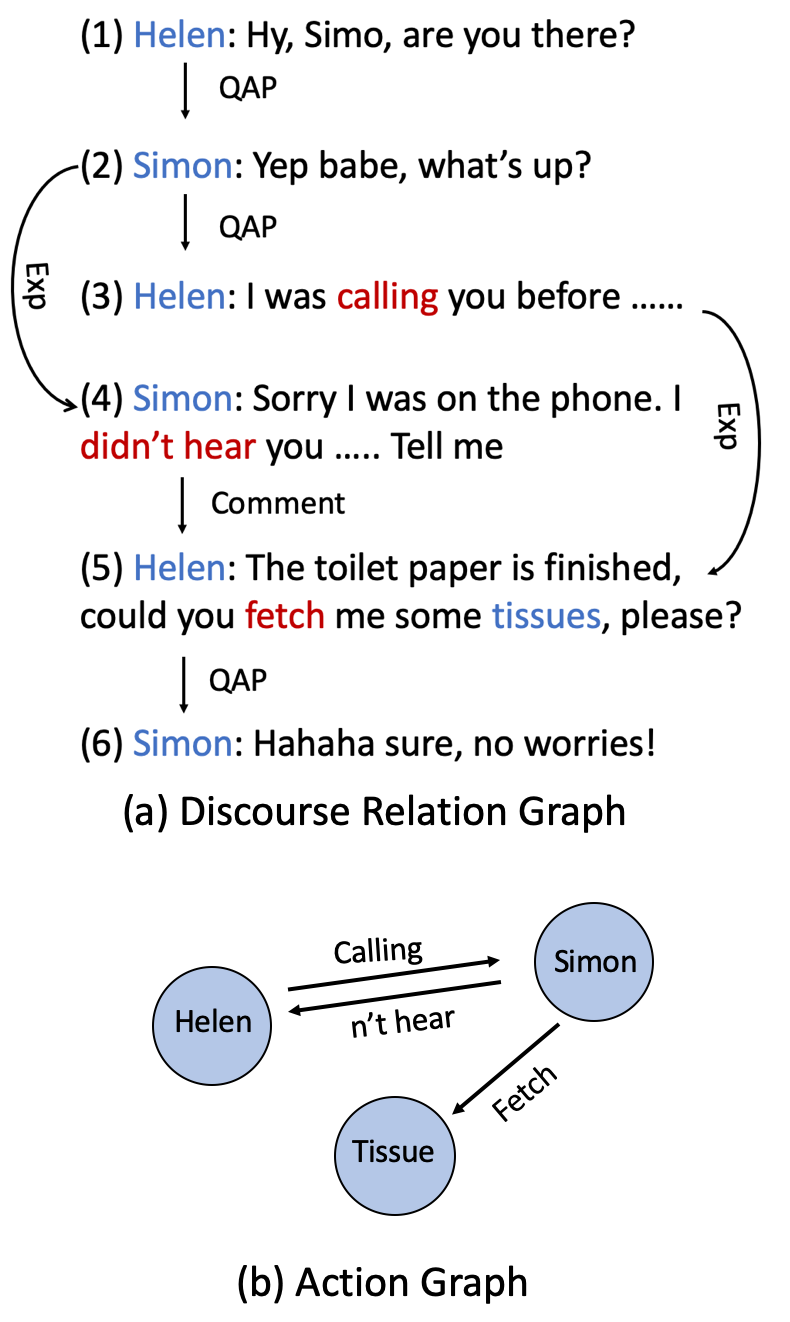}
\caption{An example of discourse relation graph (a) and action graph (b) from one conversation in SAMSum \cite{gliwa-etal-2019-samsum}. 
The annotated summary is \textit{Simon was on the phone before, so he didn’t here Helen calling. Simon will fetch Helen some tissues.} }
\label{Fig:example}
\end{figure}
Online interaction has become an indispensable component of everyday life and people are increasingly using textual conversations to exchange ideas, make plans, and share information. However, it is time-consuming to recap and grasp all the core content within every complex conversation \cite{gao2020standard,feng2020incorporating}. As a result, how to organize massive everyday interactions into natural, concise, and informative text, i.e., abstractive conversation summarization, starts to gain importance. 

Significant progress has been made on abstractive summarization for structured document via pointer generator \cite{See_2017}, reinforcement methods \cite{paulus2018a,huang-etal-2020-knowledge} and pre-trained models \cite{liu2019text,lewis-etal-2020-bart,zhang2019pegasus}. Despite the huge success, it is challenging to directly apply document models to summarize conversations, due to a set of inherent differences between conversations and documents \cite{gliwa-etal-2019-samsum}. 
First, speaker interruptions like repetitions, false-starts,  and hesitations are frequent in conversations \cite{sacks1978simplest},
and key information resides in different portions of a conversation. These unstructured properties pose challenges for models to focus on salient contents that are necessary for generating both abstractive and informative summaries. 
Second, there is more than one speaker in conversations and people interact with each other in different language styles \cite{zhu2020hierarchical}. 
% there are usually more than one speaker expressing themselves and interacting with each other in different semantic styles \cite{zhu2020hierarchical}.
%Second, there are usually multiple speakers \diyi{so far, multiple speakers have been emphasized so much -  readers might expect something in this space} in one single conversation \cite{zhu2020hierarchical}, and their roles and corresponding language use might shift with the evolution of chats. 
The complex interactions among multiple speakers make it harder for models to identify and associate speakers with correct actions so as to generate factual summaries.

In order to summarize the unstructured and complex conversations,
a growing body of research has been conducted, such as transferring document summarization methods to conversation settings \cite{shang-etal-2018-unsupervised, gliwa-etal-2019-samsum}, adopting hierarchical models \cite{10.1145/3308558.3313619, zhu2020hierarchical}, or incorporating conversation structures like topic segmentation \cite{Liu_2019, li-etal-2019-keep,chen-yang-2020-multi}, dialogue acts \cite{Goo_2018}, and conversation stages \cite{chen-yang-2020-multi}. However, current approaches still face challenges in terms of succinctness and faithfulness, as most prior studies (i) fail to explicitly model dependencies between utterances which can help identify salient portions of conversations \cite{bui-etal-2009-extracting}, and (ii) lack structured representations \cite{huang-etal-2020-knowledge} to learn the associations between speakers, actions and events. We argue that these rich linguistic structures associated with conversations are key components towards generating abstractive and factual conversation summaries. 
% \diyi{this section is okay, but not strong. how does prior work on document summarization deal with faithfulness? you do not have any related work for it now. can techniques on increasing faithfulness in document summarization be applied here? if not, why}

To this end, we present a structure-aware sequence-to-sequence model, in which we equip abstractive conversation summarization models with rich conversation structures through two types of graphs: \textbf{discourse relation graph} and \textbf{action graph}. Discourse relation graphs are constructed based on dependency-based discourse relations \cite{kirschner2012visualizing,stone2013situated,asher2016discourse,qin2017joint} between intertwined utterances, where each Elementary Discourse Unit (EDU) is one single utterance and they are linked through 16 different types of relations \cite{asher2016discourse}. As shown in Figure~\ref{Fig:example}(a), highly related utterances are linked based on discourse relations like \textit{Question Answer Pairs}, \textit{Comment} and \textit{Explanation}. Explicitly modeling these utterances relations in conversations can aid models in recognizing key content for succinct and informative summarization. Action graphs are constructed as the ``\textsc{who-doing-what}'' triplets in conversations % , which express social languages (
which express socially situated identities and activities \cite{gee2014introduction}. 
%\diyi{i did not get this term}
For instance, in Figure~\ref{Fig:example}(b), the action graph provides explicit information between \textit{Simon}, \textit{fetch}, and \textit{tissues} for the utterance \textit{it is Simon who will fetch the tissues}, making models less likely to generate summaries with wrong references (e.g., \textit{Helen will fetch the tissues}). 

To sum up, our contributions are: (1) We propose to utilize discourse relation graphs and action graphs to better encode conversations for conversation summarization. (2) We design structure-aware sequence-to-sequence models to combine these structured graphs and generate summaries with the help of a novel multi-granularity decoder. (3) We demonstrate the effectiveness of our proposed methods through experiments on a large-scale conversation summarization dataset, SAMSum \cite{gliwa-etal-2019-samsum}. (4) We further show that our structure-aware models can generalize well in new domains such as debate summarization. %  \cite{misra-etal-2015-using}.

\begin{figure*}[t]
\centering
\includegraphics[width=2.1\columnwidth]{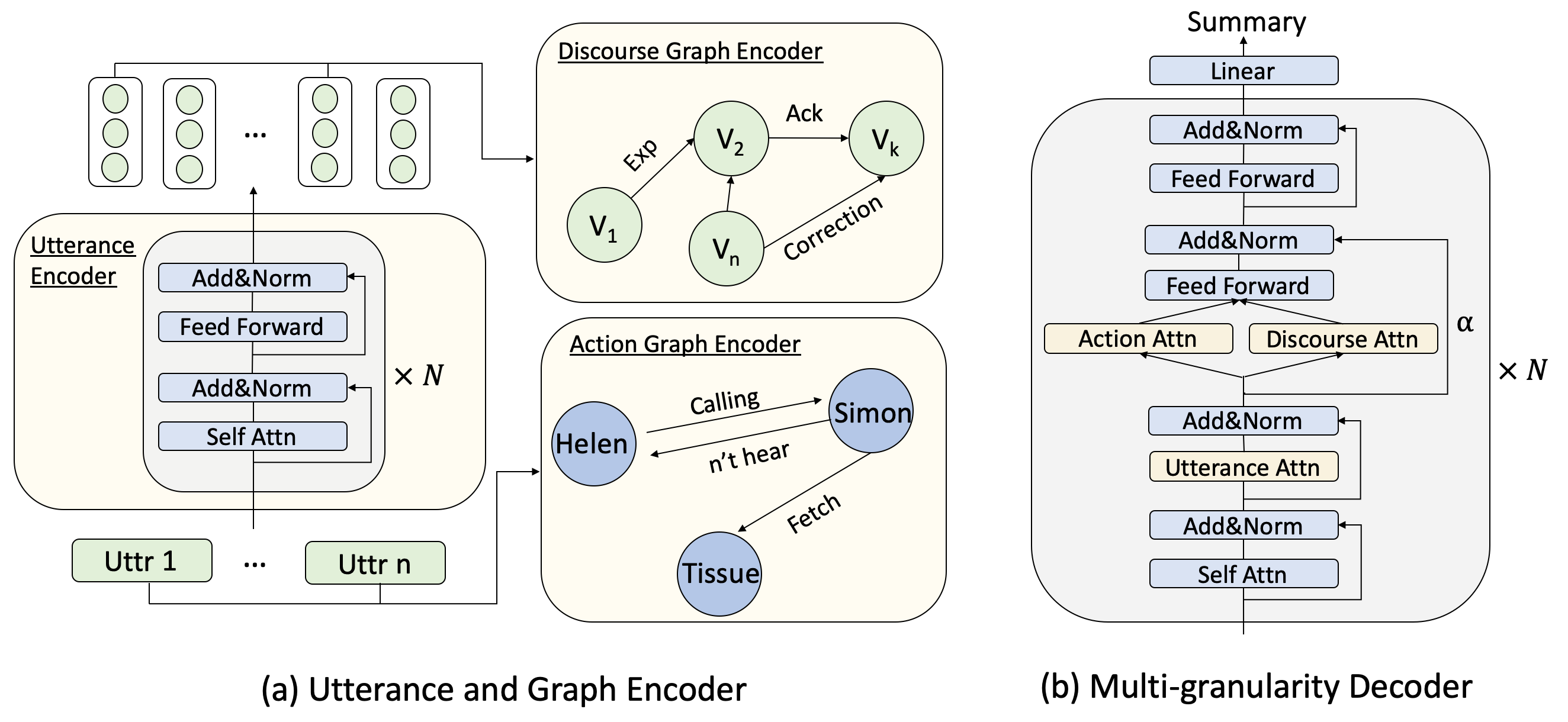}
\caption{Model architecture. Each utterance is encoded via transformer encoder; discourse relation graphs and action graphs are encoded through Graph Attention Networks (a). The multi-granularity decoder (b) then generates summaries based on all levels of encoded information including utterances, action graphs, and discourse graphs.
% \diyi{names in Figure should be consistent with the text / subsection titles}
%\diyi{the fonts need to be bigger for the figure. the color is a bit distracting - what does they mean?}
}
\label{Fig:model}
\end{figure*}

\section{Related Work}
\paragraph{Document Summarization} % Document summarization has been widely researched, especially with neural networks. 
Compared to extractive document summarization \cite{gupta2010survey,narayan2018ranking,liu2019text}, abstractive document summarization is generally considered more challenging and has received more attention. Various methods have been designed to tackle abstractive document summarization like sequence-to-sequence models \cite{rush-etal-2015-neural}, pointer generators \cite{See_2017}, reinforcement learning methods \cite{paulus2018a,huang-etal-2020-knowledge} and pre-trained models \cite{lewis-etal-2020-bart,zhang2019pegasus}. To generate faithful abstractive document summaries \cite{maynez-etal-2020-faithfulness}, graph-based models were introduced recently such as extracting entity types \cite{fernandes2018structured,fan2019using}, leveraging knowledge graphs \cite{huang-etal-2020-knowledge,Zhu2020BoostingFC} or designing extra fact correction modules \cite{dong-etal-2020-multi}. Inspired by these graph-based methods, we also construct action graphs for generating more factual conversation summaries.
% \diyi{add a sentence on how it connects to your work}
% However, most of them focused on factual knowledge triplets in single-speaker documents, and have not explicitly modeled the interactions between triples from different sources
% \diyi{this explanation is not convincing and not correct. what you used for discourse graph and actions graph are generic, and are from previous work on document relations}. 
% As a result, they could not be directly applied to summarizing multi-speaker conversations where actions from different speakers frequently interact with each other.
% \diyi{these graph based approaches can be applied, e.g., the action graph.}
    
\paragraph{Conversation Summarization} Extractive dialogue summarization \cite{Murray2005ExtractiveSO} has been studied extensively via statistical machine learning methods such as skip-chain CRFs \cite{galley2006skip}, SVM with LDA models \cite{wang2013domain}, and multi-sentence compression algorithms \cite{shang-etal-2018-unsupervised}. Such methods struggled with generating succinct, fluent, and natural summaries, especially when the key information needs to be aggregated from multiple first-person point-of-view utterances \cite{song2020ucf}. 
% There have been recent efforts on 
Abstractive conversation summarization overcomes these issues by designing hierarchical models \cite{10.1145/3308558.3313619, zhu2020hierarchical}, incorporating commonsense knowledge \cite{feng2020incorporating}, or leveraging conversational structures like dialogue acts \cite{Goo_2018}, key point sequences \cite{10.1145/3292500.3330683}, topic segments \cite{Liu_2019, li-etal-2019-keep} and stage developments \cite{chen-yang-2020-multi}. 
Some recent research has also utilized discourse relations as input features in classifiers to detect important content in conversations \cite{murray-etal-2006-incorporating,bui-etal-2009-extracting,qin2017joint}. 
However, current models still have not explicitly utilized the dependencies between different utterances, making models hard to leverage long-range dependencies and utilize these salient utterances. Moreover, 
less attention has been paid to identify the actions of different speakers and how they interact with or refer to each other, leading to unfaithful summarization with incorrect references or wrong reasoning \cite{gliwa-etal-2019-samsum}.  %\diyi{add some citation here}
% Despite several exceptions , they mainly utilized hile neglecting the actual interactions between utterances. What's more, 
% previous methods usually failed to identify the connections between speakers and behaviors, which led to unfaithful summaries with wrong references or wrong reasoning.
To fill these gaps, we propose to explicitly model actions within utterances, and relations between utterances 
% % leverage and model these interactions among utterances, speakers, and actions 
in conversations in a structured way, by using discourse relation graphs and action graphs and further combining these  through relational graph encoders  and multi-granularity decoders 
for abstractive conversation summarization.

\section{Methods}
To generate abstractive and factual summaries from unstructured conversations, we propose to model structural signals in conversations by first constructing discourse relation graphs and action graphs (Section~\ref{Sec:Graph_Construct}), and then encoding the graphs together with conversations (Section~\ref{Sec:Encoder}) as well as incorporating these different levels of information in the decoding stage through a multi-granularity decoder (Section~\ref{Sec:Decoder}) to summarize given conversations. The overall architecture is shown in Figure~\ref{Fig:model}.

\subsection{Structured Graph Construction} \label{Sec:Graph_Construct}
This section describes how to construct the discourse relation graphs and action graphs. Formally, for a given conversation $\mathbf{C} = \{\mathbf{u}_{0}, ..., \mathbf{u}_{m}\}$ with $m$ utterances, we construct \emph{discourse relation graph} $\mathcal{G}^D = (\mathbf{V}^D, \mathbf{E}^D)$, where $\mathbf{V}^D$ is the set of nodes representing Elementary Discourse Units (EDUs), and $\mathbf{E}^D$ is the adjacent matrix that describes the relations between EDUs, and  \emph{action graph} $\mathcal{G}^A = (\mathbf{V}^A, \mathbf{E}^A)$, where $\mathbf{V}^A$ is the set of nodes representing ``\textsc{who}'', ``\textsc{doing}'' and ``\textsc{what}'' arguments, and $\mathbf{E}^A$ is the adjacent matrix to link \textsc{``who-doing-what''} triples.

\paragraph{Discourse Relation Graph}
Utterances from different speakers do not occur in isolation; instead, they are related within the context of discourse \cite{murray-etal-2006-incorporating,qin2017joint}, which has been shown effective for dialogue understanding like identifying the decisions in multi-party dialogues \cite{bui-etal-2009-extracting} and detecting salient content in email conversations \cite{McKeown2007UsingQP}. Although current attention-based neural models are supposed to, or might implicitly, learn certain relations between utterances, they often struggle to focus on many informative utterances \cite{chen-yang-2020-multi,song2020ucf} and fail to address long-range dependencies \cite{xu-etal-2020-discourse}, especially when there are frequent interruptions. As a result, explicitly incorporating the discourse relations will help neural summarization models better encode the unstructured conversations and concentrate on the most salient utterances to generate more informative and less redundant summaries. 

% \cite{McKeown2007UsingQP} \cite{bui-etal-2009-extracting} \cite{murray-etal-2006-incorporating} \cite{qin2017joint}
To do so, we view each utterance as an EDU and use the discourse relation types defined in \citet{asher2016discourse}. We first pre-train a discourse parsing model \cite{shi2019deep} on a human-annotated multiparty dialogue corpus \cite{asher2016discourse}, with 0.775 F1 score on link predictions and 0.557 F1 score on relation classifications, which are comparable to the state-of-the-art results  \cite{shi2019deep}.
% \diyi{saying whether your replicated results are comparable to previous work}.
We then utilize this pre-trained parser to predict the discourse relations within conversations in our SAMSum corpus \cite{gliwa-etal-2019-samsum}. 

After predictions,  there are 138,554 edges identified in total and 8.48 edges per conversation. The distribution of these predicted discourse relation types is: Comment (19.3\%), Clarification Question (15.2\%), Elaboration (2.3\%), Acknowledgement(8.4\%), Continuation (10.1\%), Explanation (2.8\%), Conditional (0.2 \%), Question Answer Pair (21.5\%), Alternation (0.3\%), Q-Elab (2.5\%), Result (5.5\%), Background (0.4\%), Narration (0.4\%), Correction (0.4\%), Parallel (0.9\%), and Contrast (1.0\%).
% \diyi{the number of edges (not useful), and average (you shown it in Table 1) are a bit sudden, as readers might get lost on how to interpret this. }
% Based on the parsing results, 
Then for each conversation, we construct a discourse relation graph $\mathcal{G}^D = (\mathbf{V}^D, \mathbf{E}^D)$, where $\mathbf{V}^D[k]$ represents the $k$-th utterance. $\mathbf{E}^D[i][j] = r$ if there is a link from the $i$-th utterance to the $j$-th one with discourse relation $r$.

\paragraph{Action Graph} 
The ``\textit{who-doing-what}'' triples from utterances can provide explicit visualizations of speakers and their actions, the key to understanding concrete details happened in conversations
\cite{moser2001introduction,gee2014introduction,sacks1978simplest}.
% As a way of doing things with words socially together with other people \cite{sacks1978simplest}, an oral or written “utterance” in conversations has meaning only if and when it communicates a ``who'' and a ``what'' with connections (``doing'') \cite{gee2014introduction}. The ``who-doing-what'' triples provide explicit information about characters and actions, and they are fundamental for understanding the discourses in conversations \cite{moser2001introduction,gee2014introduction}. 
Simply relying on neural models to identify this information from conversations % instead of directly modeling the structured information from triples 
often fail to produce factual characterizations of concrete details happened
\cite{Cao2018FaithfulTT,huang-etal-2020-knowledge}.
To this end, we extract ``\textsc{who-doing-what}'' triples from utterances and construct  action graphs for conversation summarization \cite{chen2019incorporating,10.1145/3340531.3417466,huang-etal-2020-knowledge}.
Specifically, we first transform the first-person point-of-view utterances to its third-person point-of-view forms based on simple rules: (i) substituting first/second-person pronouns with the names of current speaker or surrounding speakers and (ii) replacing third-person pronouns based on coreference clusters in conversations detected by the Stanford CoreNLP \cite{manning-etal-2014-stanford}.
% \diyi{how well are these steps?}
For example, an utterance ``\emph{I'll bring it to you tomorrow}'' from Amanda to Jerry will be transformed into ``\emph{Amanda'll bring cakes to Jerry tomorrow}''. Then we extract ``\textsc{who-doing-what}'' (subject-predicate-object) triples from transformed conversations using the open information extraction (OpenIE) systems \footnote{\url{https://github.com/philipperemy/Stanford-OpenIE-Python}} \cite{angeli-etal-2015-leveraging}. % If two triples differ only by one argument and the rest of arguments are the same, we keep the longer triple. 
%In SAMSum corpus \cite{gliwa-etal-2019-samsum}, there are 109,810 triples extracted and the average number of triples per conversation is 6.72. \diyi{again, this information can be added to table 1}
We then construct the Action Graph $\mathcal{G}^A = (\mathbf{V}^A, \mathbf{E}^A)$ from the extracted triples by taking arguments (``\textsc{who}'', ``\textsc{doing}'', or ``\textsc{what}'' ) as nodes in $\mathbf{V}^A$, and connect them with edge $\mathbf{E}^A[i][j] = 1$ if they are adjacent in one ``\textsc{who-doing-what}'' triple. % Note that here we only have one relation type between nodes, different from discourse relation graphs.

\subsection{Encoder}  \label{Sec:Encoder}
Given a conversation and its corresponding discourse relation graph
% $\mathcal{G}^D = (\mathbf{V}^D, \mathbf{E}^D)$
and action graph, % $\mathcal{G}^A = (\mathbf{V}^A, \mathbf{E}^A)$, 
we utilize an utterance encoder and two graph encoders, to obtain its hidden representations shown in Figure~\ref{Fig:model}(a).
% we utilize neural models to encode tokens $\{x_{i,0}, ..., x_{i,l}\}$ in an utterance $\mathbf{u}_i$, nodes $\{v_0^D, ... v_m^D\}$ in discourse relation graph $\mathcal{G}^D$ and nodes $\{v_0^A, ... v_n^A\}$ in action graph ${\mathcal{G}^A}$ into hidden representation

\subsubsection{Utterance Encoder} We initialize our utterance encoder $F_U(.)$ with a pre-trained encoder, i.e., BART-base \cite{lewis-etal-2020-bart}, and encode tokens $\{x_{i,0}, ..., x_{i,l}\}$ in an utterance $\mathbf{u}_i$ into its hidden representation:
\begin{align}
    \{h^U_{i,0}, ... , h^U_{i, l}\} = F_U(\{ x_{i,0}, ... , x_{i,l} \})
\end{align}
Here we add a special token $x_{i,0}=$\textsc{<s>} at the beginning of each utterance to represent it.% the start of one utterance.

\begin{table*}[t]
\centering
\begin{tabular}{|c|c|c|c|c|c|c|}
\hline
\textbf{Dataset}                 & \textbf{Split} & \textbf{\# Conv} & \textbf{\# Participants} & \textbf{\# Turns} & \textbf{\# Discourse Edges} & \textbf{\# Action Triples} \\ \hline
\multirow{3}{*}{SAMSum} & Train & 14732   & 2.40             & 11.17     & 8.47                & 6.72               \\
                        & Val   & 818     & 2.39             & 10.83     & 8.34                & 6.48               \\
                        & Test  & 819     & 2.36             & 11.25     & 8.63                & 6.81               \\ \hline
ADSC                    & Full  & 45      & 2.00             & 7.51      & 6.51                & 37.20              \\ \hline
\end{tabular} \caption{Statistics of the used datasets, including the total number of conversations (\# Conv), the average number of participants, turns, discourse edges and action triples per conversation.} \label{Tab: data_statistics}
\end{table*}

\subsubsection{Graph Encoder}
\paragraph{Node Initialization} For \textit{discourse relation graph}, we employ the output embeddings of the special tokens $x_{i,0}$ from the utterance encoder, i.e., $h^U_{i,0}$, to initialize the $i$-th node $v_i^D$ in $\mathcal{G}^D$. We use a one-hot embedding layer to encode the relations $\mathbf{E}^D[i][j] = e^D_{i,j}$ between utterance $i$ and $j$. For \textit{action graph}, we first utilize $F_U(.)$ to encode each token in nodes $v_i^A$ and then average their output embeddings as their initial representations. %  for nodes in $\mathcal{G}^A$.

\paragraph{Structured Graph Attention Network} Based on Graph Attention Network \cite{veli2018graph}, we  utilize these relations between nodes to encode each node $v_i^D$ in $\mathcal{G}^D$ or $v_i^A$ in $\mathcal{G}^A$ through:

\begin{align*}
\alpha_{i j} &= \frac{\exp \left(\sigma\left({\mathbf{a}}^{T}\left[\mathbf{W} v_{i} \| \mathbf{W} v_{j} \| \mathbf{W_e} e_{i,j} \right]\right)\right)}{\sum_{k \in \mathcal{N}_{i}} \exp \left(\sigma \left({\mathbf{a}}^{T}\left[\mathbf{W}  v_{i} \| \mathbf{W}  v_{k}  \| \mathbf{W_e} e_{i,k} \right]\right)\right)} 
\end{align*}
\begin{align*}
    {h}_{i} &= \sigma(\sum_{j \in \mathcal{N}_{i}} \alpha_{i j} \mathbf{W} v_{j})
\end{align*}
$\mathbf{W}$, $\mathbf{W_e}$ and $\mathbf{a}$ are trainable parameters. [.$\|$.] denotes the concatenation of two vectors. $\sigma$ is the activation function, $\mathcal{N}_i$ is the set containing node-$i$'s neighbours in $\mathcal{G}$.

Through two graph encoders $F_D(., .)$ and $F_A(.,.)$, we then obtain the hidden representations of these nodes as:
\begin{align}
    \{h^D_{0}, ... , h^D_{m}\} &= F_D(\{ v^D_{0}, ... , v^D_{m} \}, \textbf{E}^D) \\
    \{h^A_{0}, ... , h^A_{n}\} &= F_A(\{ x^A_{0}, ... , x^A_{n} \}, \textbf{E}^A)
\end{align}

\subsection{Multi-Granularity Decoder}  \label{Sec:Decoder}
Different levels of encoded representations are then aggregated via our multi-granularity decoder to generate summaries as shown in Figure~\ref{Fig:model}(b). With $s-1$ previously generated tokens $y_1, ..., y_{s-1}$, our decoder $G(.)$ predicts the $l$-th token via:
\begin{align}
    \hat{y} = G(y_{1:s-1}, F_U(\mathbf{C}), F_D(\mathcal{G}^D), F_A(\mathcal{G}^A)) \\
    P(\tilde{y}_s|y_{<s}, \mathbf{C},\mathcal{G}^D,\mathcal{G}^A) = \text{Softmax}(W_p \hat{y})
\end{align}

To better incorporate the information in constructed graphs, different from the traditional pre-trained BART model \cite{lewis-etal-2020-bart}, we improve the BART transformer decoder with two extra cross attentions (Discourse Attention and Action Attention) added to each decoder layer, which attends to the encoded node representations in discourse relation graphs and action graphs. 

In each decoder layer, after performing the original cross attentions over every token in utterances $\{h^U_{i, 0:l}\}$ % that are encoded from the utterance encoder 
and getting the utterance-attended representation $x^U$, multi-granularity decoder then conducts cross attentions over nodes $\{h^D_{0:m}\}$ and $\{h^A_{0:n}\}$ that are encoded from graph encoders in parallel, to obtain the discourse-attended representation $x^D$ and action-attended representation $x^A$. These two attended vectors are then combined into a structure-aware representation $x^S$, through a feed-forward network for further forward passing in the decoder.

To alleviate the negative impact of randomly initialized graph encoders and cross attentions over graphs on pre-trained BART decoders at early stages and accelerate the learning of newly-introduced modules during training, % inspired by \citet{bachlechner2020rezero}, 
we apply ReZero \cite{bachlechner2020rezero} to the residual connection after attending to graphs in each decoder layer:
\begin{align}
    \tilde{x}^S = x^U + \alpha x^S
\end{align}
where $\alpha$ is one trainable parameter instead of a fixed value 1, which modulates updates from cross attentions over graphs.

\paragraph{Training} During training, we seek to minimize the cross entropy and use the teacher-forcing strategy \cite{bengio2015scheduled}:
\begin{align}
    \mathcal{L} = - \sum \log P(\tilde{y}_l|y_{<l}, \mathbf{C},\mathcal{G}^D,\mathcal{G}^A)
\end{align}
% We use the , i.e. the decoder takes ground-truth tokens from human generated summaries as inputs during training and takes previously predicted tokens from the decoder as inputs during inference.

\section{Experiments}
\subsection{Datasets}
We trained and evaluated our models on a conversation summarization dataset SAMSum \cite{gliwa-etal-2019-samsum} covering messenger-like conversations about daily topics, such as arranging meetings and discussing events. 
We also showed the generalizability of our models on the Argumentative Dialogue Summary Corpus (ADSC) \cite{misra-etal-2015-using}, a debate summarization corpus. The data statistics of two datasets were shown in Table~\ref{Tab: data_statistics}, with the discourse relation types distributions in the Appendix.

% [ADD one Tbale about the distribution]
% Comment (19.3\%), Clarification Question (15.2\%), Elaboration (2.3\%), Acknowledgement(8.4\%), Continuation (10.1\%), Explanation (2.8\%), Conditional (0.2 \%), Question Answer Pair (21.5\%), Alternation (0.3\%), Q-Elab (2.5\%), Result (5.5\%), Background (0.4\%), Narration (0.4\%), Correction (0.4\%), Parallel (0.9\%), and Contrast (1.0\%).

\begin{table*}[t]
\centering
\small
\begin{tabular}{|c|ccc|ccc|ccc|}
\hline
\multirow{2}{*}{\textbf{Model}}         & \multicolumn{3}{c|}{\textbf{ ROUGE-1}} & \multicolumn{3}{c|}{\textbf{ ROUGE-2}} & \multicolumn{3}{c|}{\textbf{ ROUGE-L}}\\ \cline{2-10}
                                    & F       & P       & R       & F       & P       & R       & F       & P       & R       \\ \hline
Pointer Generator \cite{See_2017}                    & 40.08   & -       & -       & 15.28   & -       & -       & 36.63   & -       & -       \\
Transformer \cite{vaswani2017attention}              & 37.27   & -       & -       & 10.76   & -       & -       & 32.73   & -       & -       \\
D-HGN \cite{feng2020incorporating}  & 42.03   & -       & -       & 18.07   & -       & -       & 39.56   & -       & -  \\   
%\multirow{1}{*}{Topic BART }                   & 45.36   & 50.78   & 45.61   & 21.83   & 24.73   & 22.06   & 44.56   & 49.86   & 43.96   \\
%\multirow{1}{*}{Stage BART }                    & 45.38   &50.43   & 45.93   & 22.23   & 24.66   & 22.48   & 44.68   & 49.53   & 44.35   \\  %\hline
\multirow{1}{*}{Multi-view Seq2Seq \cite{chen-yang-2020-multi}}                    & 45.56   & 52.13   & 44.68   & 22.30   & 25.58   & 22.03   & 44.70   & 50.82   & 43.29   \\ \hline 
\multirow{1}{*}{BART \cite{lewis-etal-2020-bart}} & 45.15   & 49.58   & 45.97   & 21.66   & 23.95   & 22.16   & 44.46   & 48.92   & 44.26   \\ \hline
\multirow{1}{*}{S-BART w. Discourse  $\dag$  } & 45.89   & 51.34   & 45.87   & 22.50   & 25.26   & 22.33   & 44.83   & 49.93   & 44.17   \\
\multirow{1}{*}{S-BART w. Action $\dag$  } & 45.67   & 50.25   & 46.44   & 22.39   & 24.70   & 22.96   & 44.86   & 49.29   & 44.75   \\
\multirow{1}{*}{S-BART w. Discourse\&Action  $\dag$ } & \textbf{46.07}   & 51.13   & 46.24   & \textbf{22.60}   & 25.11   & 22.81   & \textbf{45.00}   & 49.82   & 44.47   \\ \hline 
 
\end{tabular} 
\caption{ROUGE-1,  ROUGE-2 and  ROUGE-L scores for different models on the SAMSum Corpus test set. Results are averaged over three random runs. $\dag$ means our methods. 
We performed Pitman's permutation test \cite{dror2018hitchhiker} and found that \emph{S-BART w. Discourse\& Action} significantly outperformed the base \emph{BART} (p < 0.05).
% \diyi{the results might look a bit "incremental" to reviewers. it might make sense to emphasize the significance of this improvement, and human eval (if there are big gains), or other dimensions that you optimized... though error analyses are not so necessary, it might be good to manually go through some and talk about how graph helps reduce errors, etc. }
} \label{Tab: Main_results}
\end{table*}

\begin{table*}[t]
\centering
\small
\begin{tabular}{|c|ccc|ccc|ccc|}
\hline
\multirow{2}{*}{\textbf{Model}}         & \multicolumn{3}{c|}{\textbf{ ROUGE-1}} & \multicolumn{3}{c|}{\textbf{ ROUGE-2}} & \multicolumn{3}{c|}{\textbf{ ROUGE-L}}\\ \cline{2-10}
                                    & F       & P       & R       & F       & P       & R       & F       & P       & R       \\ \hline 
\multirow{1}{*}{ BART \cite{lewis-etal-2020-bart}} & 20.90   & 51.71   & 13.53   & 5.04   & 12.46   & 3.24   & 21.23   & 56.29   & 13.54   \\
 \hline  \hline 
\multirow{1}{*}{S-BART w. Discourse  $\dag$  } & 22.42   & 54.13   & 14.60   & 5.58   & 13.83   & 3.61   & 22.16   & 51.88  & 14.45   \\
\multirow{1}{*}{S-BART w. Action  $\dag$  } & 30.91   & 85.42   & 19.12   & 20.64   & 56.31   & 12.78   & 35.30   & 85.51   & 22.58   \\
\multirow{1}{*}{S-BART w. Discourse\&Action $\dag$ } & \textbf{34.74}   & 84.99   & 22.20   & \textbf{23.86}   & 58.08   & 15.24   & \textbf{38.69}   & 83.81   & 25.51   \\ \hline 
 
\end{tabular} 
\caption{ROUGE-1, ROUGE-2 and ROUGE-L scores on the out-of-domain ADSC corpus using different models trained on SAMSum Corpus. $\dag$ means our methods.} \label{Tab: Zero_shot}
\end{table*}

\subsection{Baselines}
We compare our methods with several baselines:
% \textbf{Pointer Generator} \cite{See_2017}: We followed the settings in \citet{gliwa-etal-2019-samsum} and used special tokens to separate each utterance.

% \textbf{Transformer} \cite{vaswani2017attention}: We trained transformer seq2seq models following the OpenNMT \cite{klein2017opennmt}.
    
% \textbf{D-HGN} \cite{feng2020incorporating} incorporated commonsense knowledge from ConceptNet \cite{10.1023/B:BTTJ.0000047600.45421.6d} for dialogue summarization.
    
% \textbf{BART} \cite{lewis-etal-2020-bart}: We utilized BART
    % \footnote{\url{https://huggingface.co/transformers/model_doc/bart.html}}, and separated utterances  by a special token.
    
% \textbf{Multi-View Seq2Seq} \cite{chen-yang-2020-multi} % was based on BART \cite{lewis-etal-2020-bart}, and 
% utilized topic and stage views on top of BART for summarizing conversations. % Here we implemented it based on BART-base models.

\begin{itemize}\setlength\itemsep{1em}
    \item \textbf{Pointer Generator} \cite{See_2017}: We followed the settings in \citet{gliwa-etal-2019-samsum} and used special tokens to separate each utterance.
    \item \textbf{Transformer} \cite{vaswani2017attention}: We trained transformer seq2seq models following the OpenNMT \cite{klein2017opennmt}.
    \item \textbf{D-HGN} \cite{feng2020incorporating} incorporated commonsense knowledge from ConceptNet \cite{10.1023/B:BTTJ.0000047600.45421.6d} for dialogue summarization.
    \item \textbf{BART} \cite{lewis-etal-2020-bart}: We utilized BART
    \footnote{The version on 10/7 in \url{https://huggingface.co/transformers/model_doc/bart.html}}, and separated utterances  by a special token.
    \item \textbf{Multi-View Seq2Seq} \cite{chen-yang-2020-multi} % was based on BART \cite{lewis-etal-2020-bart}, and 
    utilized topic and stage views on top of BART for summarizing conversations. Here we implemented it based on BART-base models.
\end{itemize}

\subsection{Implementation Details}
%We pre-trained a deep sequential model \cite{shi2019deep} on STAC Corpus (1,062 dialogues) \cite{asher2016discourse} with default settings \footnote{\url{https://github.com/shizhouxing/DialogueDiscourseParsing}} to get the link prediction and relation classification models to label discourse relations in SAMSum and ADSC corpus. 
We used the BART-base model to initialize our sequence-to-sequence model for training in all experiments. For parameters in the original BART encoder/decoder, we followed the default settings and set the learning rate 3e-5 with 120 warm-up steps. For graph encoders, we set the number of hidden dimensions as 768, the number of attention heads as 2, the number of layers as 2, and the dropout rate as 0.2. For graph cross attentions added to BART decoder layers, we set the number of attention heads as 2. The weights $\alpha$ in ReZero residual connections were initialized with 1. The learning rate for parameters in newly added modules was 3e-4 with 60 warm-up steps. All experiments were performed on GeForce RTX 2080Ti (11GB memory).

\subsection{Results on In-Domain Corpus}
\begin{table}[t]
\small
\centering
\begin{tabular}{|c|c|c|c|}
\hline
\multirow{1}{*}{\textbf{Models}}       & \multicolumn{1}{c|}{\textbf{Fac.}}  & \multicolumn{1}{c|}{\textbf{Suc.}} & \multicolumn{1}{c|}{\textbf{Inf.}} \\ \hline
    \multirow{1}{*}{Ground Truth} & \textbf{4.29}     & 4.40      & \textbf{4.06}       \\ \hline 
\multirow{1}{*}{BART} &  3.90      &  4.13     & 3.74      \\  \hline
\multirow{1}{*}{S-BART w. Discourse} & 4.11      & \textbf{4.42}     & 3.98   \\ 
\multirow{1}{*}{S-BART w. Action} &  4.17    &  4.29      & 3.95   \\ 
\multirow{1}{*}{S-BART w. Discourse\&Action} & 4.19  & 4.41      & 3.91      \\\hline 
 
\end{tabular} 
\caption{Human evaluation on \textbf{Fac}tualness, \textbf{Suc}cinctness, \textbf{Inf}ormativeness. All model variants of \textit{S-BART} received significantly higher ratings than \textit{BART} (student t-test, p < 0.05). }\label{Tab: Human}
\end{table}
\paragraph{Automatic Evaluation}
We evaluated all the models with the widely used automatic metric, ROUGE scores \cite{lin2004automatic} \footnote{We followed fairseq and used \url{https://github.com/pltrdy/rouge} to calculate ROUGE scores. Note that different tools may result in different ROUGE scores.}, and reported ROUGE-1, ROUGE-2, and ROUGE-L %. The results on the SAMSum test set were shown
in Table~\ref{Tab: Main_results}. 
We found that, 
% The first section displayed the previous state-of-the-art models.
compared to simple sequence-to-sequence models (\textit{Pointer Generator} and \textit{Transformer}), incorporating extra information such as commonsense knowledge from ConceptNet (\textit{D-HGN}) increased the ROUGE metrics. When equipped with pre-trained models and simple conversation structures such as topics and conversation stages, \textit{Multi-View Seq2Seq} boosted ROUGE scores.
% The results of our base model (\textit{ BART}) and proposed methods (\textit{S-BART}) were shown in the second section. Compared to BART with conversation views, \textit{BART} showed lower ROUGE scores as it failed to model rich conversation structures. %We tested our structure-aware BART (\textit{S-BART w. Discourse/Action}) within two ReZero settings: (i) initializing $\alpha$ from 0, (ii) initializing $\alpha$ from 1. 
Incorporating discourse relation graphs or action graphs helped the performances of summarization, suggesting the effectiveness of explicitly modeling relations between utterances and the associations between speakers and actions within utterances. Combining two different structured graphs produced better ROUGE scores compared to previous state-of-the-art methods and our base models, with an increase of 2.0\% on ROUGE-1, 4.3\% on ROUGE-2, and 1.2\% on ROUGE-L compared to our base model, \textit{BART}. This indicates that, % which further indicated that 
our structure-aware models with discourse and action graphs could help abstractive conversation summarization, and these two graphs complemented each other in generating better summaries.

\paragraph{Human Evaluation} 
We conducted human evaluation to qualitatively evaluate the generated summaries. Specifically, we asked 
% native speakers of English on 
annotators from Amazon Mechanical Turk  to score a set of randomly sampled 100 generated summaries from ground-truth, BART and our structured models,  % \textit{Human Annotations,  BART, S-BART w. Discourse, S-BART w. Action and S-BART w. Discourse\&Action} 
using  a Likert scale from 1 (worst) to 5 (best) in terms of \textbf{factualness} (e.g., associates actions with the right actors)  , \textbf{succinctness} (e.g., does not contain redundant information), and \textbf{informativeness} (e.g., covers the most important content) \cite{feng2020incorporating, huang-etal-2020-knowledge}.
To increase annotation quality, we required turkers to have a 98\%
approval rate and at least 10,000 approved tasks for their previous work. Each message was rated by three workers.
The scores for each summary were averaged. The Intra-Class Correlation was 0.543, showing moderate agreement \cite{koo2016guideline}.

As shown in Table~\ref{Tab: Human}, \textit{S-BART} that utilized structured information from discourse relation graphs and action graphs generated significantly better summaries with respect to factualness, succinctness, and informativeness.
This might because that the incorporation of structured information such as discourse relations helped \textit{S-BART} to recognize the salient parts in conversations, and thus improve the succinctness and informativeness over  \textit{BART}. % (e.g., \textit{S-BART w. Discourse} achieved similar scores as \textit{Human Annotations}) compared to .
Modeling the connections between speakers and actions greatly helped generate more factual summaries than the baselines, e.g., with an increase of 0.27 from \textit{BART} to \textit{S-BART w. Action}.

\subsection{Results on Out-Of-Domain Corpus}   
To investigate the generalizability of our structure-aware models, we then tested the \textit{S-BART} model trained on SAMSum corpus directly on the debate summarization domain (ADSC Corpus \cite{misra-etal-2015-using}) in a zero-shot setting. 
Besides the differences in topics, % debate conversations are different compared to daily chats in conversations, 
% , compared to daily chats, 
utterances in debate conversations were generally longer and include more action triples (37.20 vs 6.81 as shown in Table \ref{Tab: data_statistics}) and fewer participants.
The distribution of discourse relation types
also differed a lot across different domains\footnote{The detailed distributions were shown in the Appendix.} (e.g.,  more \textit{Contrast} in debates (19.5\%) than in daily conversations (1.0\%)). %These differences made the debate conversations significantly different from daily chats. 

As shown in Table~\ref{Tab: Zero_shot},  our single graph models \textit{S-BART w. Discourse} and \textit{S-BART w. Action} boosted ROUGE scores compared to \textit{BART},
suggesting that utilizing structures can also increase the generalizability of conversation summarization methods. However, contrary to in-domain results in Table~\ref{Tab: Main_results}, action graphs led to much more gains than discourse graphs.
This indicated that when domain shifts, action triples were most robust in terms of zero-shot setups; differences in discourse relation distributions could limit such generalization. 
% indicating that the discrepancy of discourse relation distributions limited the zero-shot abilities when domain shifted, while action triples kept more robust with domain variations. 
Consistent with in-domain scenarios, our \textit{S-BART w. Discourse\&Action} achieved better results, with an increase of 66.2\% on ROUGE-1, 373.4\% on ROUGE-2, and 82.2\% on ROUGE-L over \textit{BART}.

\subsection{Ablation Studies} 
This part conducted ablation studies to show the effectiveness of structured graphs in our \textit{S-BART}.
\begin{table}[t]
\centering
\small
\begin{tabular}{|c|c|c|c|}
\hline
\multirow{1}{*}{\textbf{Graph Types}}         & \multicolumn{1}{c|}{\textbf{ R-1}} & \multicolumn{1}{c|}{\textbf{ R-2}} & \multicolumn{1}{c|}{\textbf{ R-L}}\\ \hline
%\multirow{1}{*}{Discourse Graph $\alpha = 0$ } & \textbf{45.40}    & \textbf{21.96}      & \textbf{44.56}       \\
%\multirow{1}{*}{Random Graph $\alpha = 0$ } & 45.36      & 21.76     & 44.23      \\ \hline
\multirow{1}{*}{S-BART w. Discourse Graph } & \textbf{45.89}      & \textbf{22.50}      & \textbf{44.83}     \\
\multirow{1}{*}{S-BART w. Random Graph } & 45.28      & 21.80      & 44.30      \\\hline 
 
\end{tabular} 
\caption{ROUGE-1,  ROUGE-2 and ROUGE-L scores of S-BART with either the constructed discourse relation graphs or random graphs. Results are averaged over three random runs.} \label{Tab: Ablation_1}
\end{table}

\begin{table}[t]
\centering
\small
\begin{tabular}{|c|c|c|c|}
\hline
\multirow{1}{*}{\textbf{Combination Strategy}}         & \multicolumn{1}{c|}{\textbf{ R-1}} & \multicolumn{1}{c|}{\textbf{ R-2}} & \multicolumn{1}{c|}{\textbf{ R-L}}\\ \hline
\multirow{1}{*}{Parallel} & \textbf{46.07}   & \textbf{22.60}     & \textbf{45.00}       \\ \hline
\multirow{1}{*}{Sequential (discourse, action) } & 45.40      & 22.14    & 44.67      \\ 
\multirow{1}{*}{Sequential (action, discourse)} & 45.62      & 22.41     & 44.62    \\ \hline
\end{tabular} 
\caption{ROUGE-1,  ROUGE-2 and ROUGE-L scores of S-BART models using different ways to combine discourse relation graphs and action graphs. Results are averaged over three random runs.} \label{Tab: Ablation_2}
\end{table}

\begin{figure}[t]
\centering
\includegraphics[width=1.0\columnwidth]{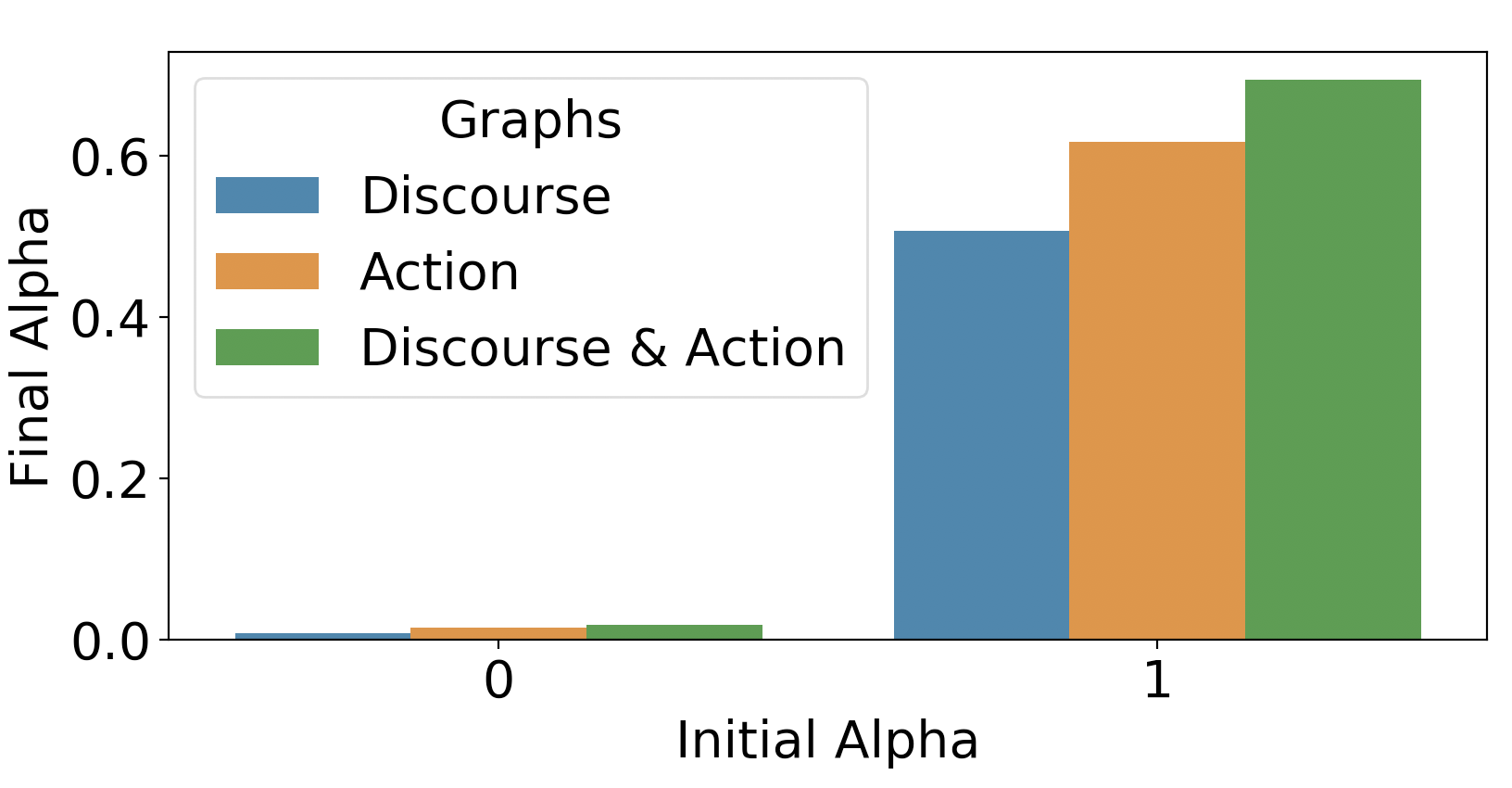}
\caption{Averaged $\alpha$ over decoder layers in the trained S-BART models using different graphs}
\label{Fig:alpha_weight}
\end{figure}

\paragraph{The Quality of Discourse Relation Graphs}
We showed how the quality of discourse relation graphs affected the performances of conversation summarization in Table~\ref{Tab: Ablation_1}. 
Specifically, we compared the ROUGE scores of S-BART using our constructed discourse relation graphs (\textit{S-BART w. Discourse Graph}) and S-BART using randomly generated discourse relation graphs \textit{S-BART w. Random Graph} where both connections between nodes and relation types were randomized.  
The number of edges in two graphs was kept the same. 
We found that S-BART with our discourse graphs outperformed models with random graphs, indicating the effectiveness of the constructed discourse relation graphs and the importance of their qualities. %  Also, such observations suggested that incorporating discourse graphs with the higher quality might bring in more performance gains, which revealed the need for better frameworks or models on conversation discourse analyses. 
% \diyi{this comparison is a bit weak, as it only compares random graph - one condition}

\paragraph{Different Ways to Combine Graphs}
We experimented with different ways to combine discourse relation graphs and action graphs in our \textit{S-BART w. Discourse \& Action}, and presented  the results in Table~\ref{Tab: Ablation_2}.
Here, \textbf{parallel} strategy performed cross attentions on different graphs separately and then combined the attended results with feed-forward networks as discussed in Section~\ref{Sec:Decoder};
\textbf{sequential} strategy performed cross attentions on two graphs in a specific order (from discourse relation graphs to actions graphs, or vice versa). 
We found that the parallel strategy showed better performances and the sequential ones did not introduce gains compared to S-BART with single graphs. 
This demonstrates that discourse relation graphs and action graphs were both important and provided different signals for abstractive conversation summarization.

\paragraph{Visualizing ReZero Weights} We further tested our structure-aware BART with two ReZero settings: (i) initializing $\alpha$ from 0, (ii) initializing $\alpha$ from 1, and found initializing $\alpha$ from 1 would bring in more performance gains (see Appendix). We then visualized the average $\alpha$ over different decoder layers after training in Figure~\ref{Fig:alpha_weight}, and observed that (i) when $\alpha$ was initialized with 1, the final $\alpha$ was much larger than the setting where $\alpha$ was initialized with 0, which might because randomly initialized modules barely received supervisions at early stages and therefore contributes less to \textit{BART}. % \textit{S-BART} with higher $\alpha$ outperforming \textit{S-BART} with $\alpha$ that was close to 0 also indicated that our introduced graph modules did contribute to the better summary generation. 
(ii) Compared to discourse graphs, action graphs received higher $\alpha$ weights after training in both initializing settings, suggesting that the information from structured action graphs might be harder for the end-to-end \textit{BART} models to capture. (iii) Utilizing both graphs spontaneously led to higher ReZero weights, further validating the effectiveness of combining discourse relation graphs and action graphs and their complementary properties. % displayed that they could complement each other.

\begin{table}[t]
%small
\centering
\begin{tabular}{|c|c|c|c|}
\hline
\multirow{1}{*}{\textbf{Conversations}} & \multicolumn{1}{c|}{\textbf{\# Num}} & \multicolumn{1}{c|}{\textbf{\# Dis.}}      & \multicolumn{1}{c|}{\textbf{\# Act.}}   \\ \hline
\multirow{1}{*}{Test Set}  & 819 &  8.63   & 6.81               \\ \hline 
\multirow{1}{*}{Similar} & 373 & 8.31   &  6.36                 \\ 
\multirow{1}{*}{Increase} & 208 & 9.13   & 7.40                \\ 
\multirow{1}{*}{Challenging} & 160 & \textbf{9.58}   &  \textbf{7.85}          \\ \hline 

\end{tabular} 
\caption{The total number of examples, average number of \textbf{Dis}course edges and \textbf{Act}ion triples in different set of conversations in the SAMSUM test set.% \diyi{how many data points here? what are the percentage}
}  \label{Tab: Case}
\end{table}

\subsection{Error Analyses}
To inspect when our summarization models could help the conversations summarization, we visualized the average number of discourse edges and the average number of action triples in three sets of conversations in Table~\ref{Tab: Case}: (i) \textbf{Similar}: examples where \textit{S-BART} generated similar ROUGE scores (the differences were less than 0.1) compared to \textit{BART}; (ii) \textbf{Increase}: examples where \textit{S-BART} resulted in higher ROUGE scores (the differences were larger than 1.0)  compared to \textit{BART}; (iii) \textbf{Challenging}: examples where both \textit{S-BART} and \textit{BART} showed low ROUGE scores (ROUGE-1 < 20.0, ROUGE-2 < 10.0, ROUGE-L < 10.0). 
% \diyi{what are the specific range/gap/delta when defining similar, increase or low?}

When the structures in conversations were simpler (fewer discourse edges and fewer action triples than the average), \textit{BART} showed similar performance as  \textit{S-BART}. As the structures of conversations become more complex with more discourse relations and more action mentions, \textit{S-BART} outperformed \textit{BART} as it explicitly incorporated these structured graphs. However, both \textit{BART} and  \textit{S-BART} struggled when % generated summaries with lower ROUGE scores when
there were much more interactions beyond certain thresholds, calling for better mechanisms to model structures in conversations for generating better summaries. % summarizing complicated conversations.

\section{Conclusion} 
In this work, we introduced a structure-aware sequence-to-sequence model for abstractive conversation summarization by incorporating discourse relations between utterances, and the connections between speakers and actions within utterances.  Experiments and ablation studies on SAMSum corpus showed the effectiveness of these structured graphs in aiding the task of conversation summarization via both quantitative and qualitative evaluation metrics. Results in zero-shot settings on ADCS Corpus further demonstrated the generalizability of our structure-aware models. 
In the future, we plan to extend our current conversation summarization models for various application domains such as emails, debates, and podcasts, and in conversations that might involve longer utterances and more participants in an unsynchronized way.
% For future directions, we would explore conversation in different domains.
% multiple domains. 

% \section{Ethical Considerations}
% The conversation summarization models developed in our work would be beneficial for a wide range of populations via automatically providing key points of chats or discussions from phone calls, remote video conferences, or text conversations to help people quickly recap previous interactions. %But the system might encourage people to escape from necessary communications, which might harm other attendees. 
% Failure models like failing to cover all the important content or generating unfaithful summaries could mislead users from capturing precise information in previous conversations, thus leading to wrong impressions. Furthermore, due to the subjective nature and goal of mimicking behaviors from dataset, the conversation summarization system might be susceptible to implicitly encode and reflect underlying biases in generated summaries, e.g., gender bias, personal point-of-view bias, or biases on quiet speakers, which brings up challenges for models to avoid these biases when they are applied to real-world settings.

\section*{Acknowledgment }
We would like to thank the anonymous reviewers
for their helpful comments, and the members of Georgia Tech SALT group for their feedback. This work is supported in part by grants from Google, Amazon and Salesforce.

\bibliography{anthology}
\bibliographystyle{acl_natbib}

\appendix

\begin{table*}[t]
\centering
\small
\begin{tabular}{|c|ccc|ccc|ccc|}
\hline
\multirow{2}{*}{\textbf{Model}}         & \multicolumn{3}{c|}{\textbf{ ROUGE-1}} & \multicolumn{3}{c|}{\textbf{ ROUGE-2}} & \multicolumn{3}{c|}{\textbf{ ROUGE-L}}\\ \cline{2-10}
                                    & F       & P       & R       & F       & P       & R       & F       & P       & R       \\ \hline
\multirow{1}{*}{BART \cite{lewis-etal-2020-bart}} & 45.15   & 49.58   & 45.97   & 21.66   & 23.95   & 22.16   & 44.46   & 48.92   & 44.26   \\ \hline 

\multirow{1}{*}{S-BART w. Discourse $\alpha = 0$ $\dag$} & 45.40  & 50.22   & 45.86   & 21.96   & 24.49   & 22.25   & 44.56   & 49.32   & 44.13   \\
\multirow{1}{*}{S-BART w. Action $\alpha = 0$ $\dag$} & 45.47   & 50.82   & 45.42   & 22.23   & 24.96   & 22.34   & 44.55   & 49.69   & 43.75   \\
\multirow{1}{*}{S-BART w. Discourse\&Action $\alpha = 0$ $\dag$} & \textbf{45.59}   & 51.47   & 45.09   & \textbf{22.42}   & 25.51   & 22.27   & \textbf{44.67}   & 50.24  & 43.52   \\ \hline

\multirow{1}{*}{S-BART w. Discourse $\alpha = 1$  $\dag$  } & 45.89   & 51.34   & 45.87   & 22.50   & 25.26   & 22.33   & 44.83   & 49.93   & 44.17   \\
\multirow{1}{*}{S-BART w. Action $\alpha = 1$ $\dag$  } & 45.67   & 50.25   & 46.44   & 22.39   & 24.70   & 22.96   & 44.86   & 49.29   & 44.75   \\
\multirow{1}{*}{S-BART w. Discourse\&Action  $\alpha = 1$ $\dag$ } & \textbf{46.07}   & 51.13   & 46.24   & \textbf{22.60}   & 25.11   & 22.81   & \textbf{45.00}   & 49.82   & 44.47   \\ \hline 
 
\end{tabular} 
\caption{Results on SAMSum Corpus. ROUGE-1,  ROUGE-2 and  ROUGE-L scores for different models on the test set. Results are averaged over three random runs. $\dag$ means our methods.} \label{Tab: ReZero Results}
\end{table*}

\section{Discourse Relation Distributions}
We pre-trained a deep sequential model \cite{shi2019deep} on STAC Corpus (1,062 dialogues) \cite{asher2016discourse} with default settings \footnote{\url{https://github.com/shizhouxing/DialogueDiscourseParsing}} to get the link prediction and relation classification models to label discourse relations in SAMSum and ADSC corpus. The distribution of the relation types in two datasets were shown in Table~\ref{Tab: Distribution}. The major discourse relations in daily conversations are Comment, Clarification and QA pairs, while the main discourse relations in debate are Comment, Contrast, Clarification and QA pairs.

\begin{table}[t]
\centering
\begin{tabular}{|c|cc|}
\hline
\textbf{Discourse Type} & \textbf{SAMSum} & \textbf{ADSC}   \\ \hline
Comment        & 19.3\% & 42.7\% \\
Clarification  & 15.2\% & 13.3\% \\
Elaboration    & 2.3\%  & 0.1\%  \\
Acknowlegement & 8.4\%  & 0.9\%  \\
Explanation    & 2.8\%  & 0.3\%  \\
Conditional    & 0.2\%  & 0\%    \\
QA pair        & 21.5\% & 12.3\% \\
Alternation    & 0.3\%  & 0.6\%  \\
Result         & 5.5\%  & 0.2\%  \\
Backgraound    & 0.4\%  & 0\%    \\
Narration      & 0.4\%  & 0\%    \\
Correction     & 0.4\%  & 1.1\%  \\
Continuation   & 0.9\%  & 7.5\%  \\
Q-Elab         & 2.5\%  & 0\%    \\
Parallel       & 0.9\%  & 0\%    \\
Contrast       & 1.0\%  & 19.5\% \\ \hline

\end{tabular}
\caption{The distribution of predicted discourse relation types on SAMSum Corpus and ADSC Corpus.}\label{Tab: Distribution}
\end{table}

\section{Impact of Different ReZero Weight Initializations}
We tested our structure-aware BART (\textit{S-BART w. Discourse/Action}) within two ReZero settings: (i) initializing $\alpha$ from 0, (ii) initializing $\alpha$ from 1.  And the results were shown in Table~\ref{Tab: ReZero Results}. \textit{S-BART} with 1 as the initialized ReZero weight outperformed that with 0 under under all graph settings, suggesting utilizing more information from graphs would bring in more performance boosts.

\end{document}